\documentclass{article}

\usepackage[preprint, nonatbib]{neurips_2020}

\usepackage[utf8]{inputenc} 
\usepackage[T1]{fontenc}    
\usepackage{hyperref}       
\usepackage{url}            
\usepackage{booktabs}       
\usepackage{amsfonts}       
\usepackage{nicefrac}       
\usepackage{microtype}      
\usepackage{graphicx}
\usepackage{listings}
\usepackage{xcolor}
\usepackage{subfig}
\usepackage{amsmath}
\usepackage{wrapfig}
\usepackage{color}

\title{Elastic-InfoGAN: Unsupervised Disentangled Representation Learning in Class-Imbalanced Data}

\author{%
  Utkarsh Ojha$^{1}$~~~~~Krishna Kumar Singh$^{1,2}$~~~~~Cho-Jui Hsieh$^{3}$~~~~~Yong Jae Lee$^{1}$\\
  \\\\
  $^{1}$UC Davis \hspace{10mm} $^{2}$Adobe Research \hspace{10mm} $^{3}$UCLA \\\\
  \url{utkarshojha.github.io/elastic-infogan/}\\
   
}
\begin{document}

\maketitle


\begin{abstract}
   We propose a novel unsupervised generative model that learns to disentangle object identity from other low-level aspects in class-imbalanced data. We first investigate the issues surrounding the assumptions about uniformity made by InfoGAN~\cite{chen-nips16}, and demonstrate its ineffectiveness to properly disentangle object identity in imbalanced data. Our key idea is to make the discovery of the discrete latent factor of variation invariant to identity-preserving transformations in real images, and use that as a signal to learn the appropriate latent distribution representing object identity. Experiments on both artificial (MNIST, 3D cars, 3D chairs, ShapeNet) and real-world (YouTube-Faces) imbalanced datasets demonstrate the effectiveness of our method in disentangling object identity as a latent factor of variation. 
\end{abstract}


\section{Introduction}

Generative models aim to model the true data distribution, so that \emph{fake} samples that seemingly belong to the modeled distribution can be generated \cite{ackley-cs1985,rabiner-ieee1989,blei-jmlr2003}.  Recent deep neural network based models such as Generative Adversarial Networks \cite{goodfellow-nips2014,salimans-nips16,radford-iclr16} and Variational Autoencoders \cite{kingma-iclr2014,higgins-iclr17} have led to promising results in generating realistic samples for high-dimensional and complex data such as images.  More advanced models show how to discover \emph{disentangled} (factorized) representations \cite{yan-eccv16,chen-nips16,tran-cvpr2017,hu-cvpr18,singh-cvpr2019}, in which different latent dimensions can be made to represent independent factors of variation (e.g., pose, identity) in the data (e.g., human faces). 

\begin{figure*}[t!]
    \centering
    \includegraphics[width=0.85\textwidth]{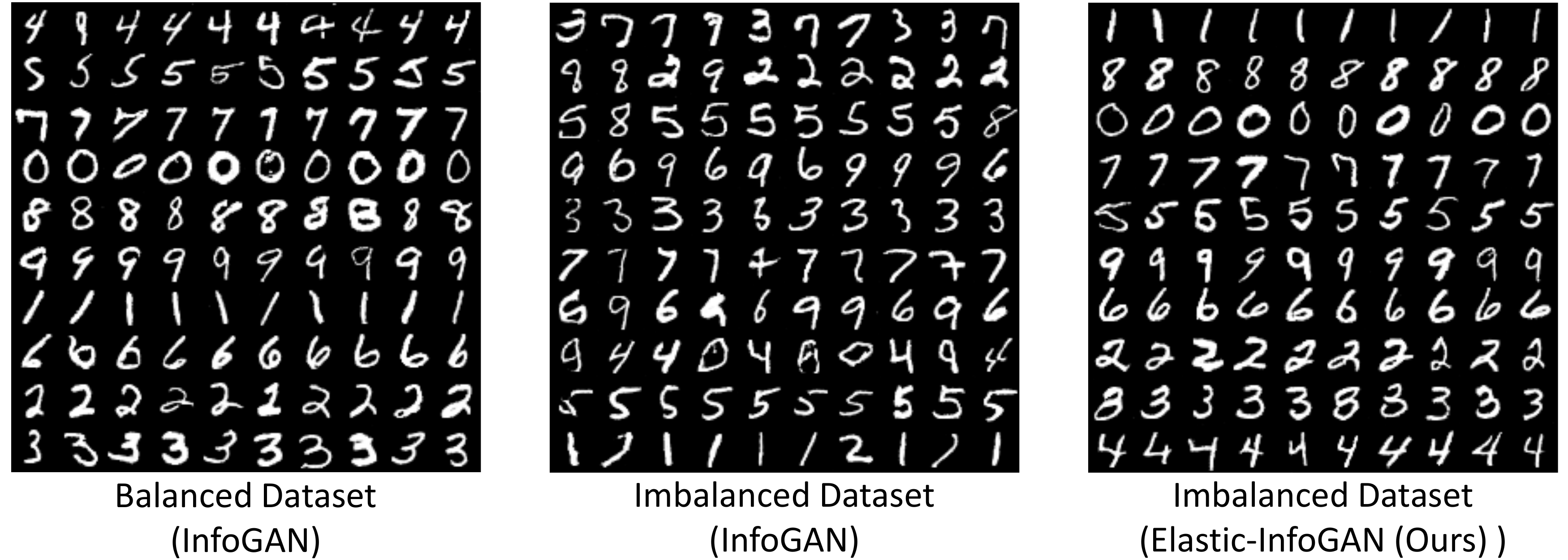}
    \caption{(\textbf{Left} \& \textbf{Center}) Samples generated with an InfoGAN model learned with a fixed uniform categorical distribution $Cat(K = 10, p = 0.1)$ on balanced and imbalanced data, respectively.  Each row corresponds to a different learned latent category.  (\textbf{Right}) Samples generated with Elastic-InfoGAN using its automatically learned latent categorical distribution. Although InfoGAN discovers digit identities in the balanced data, it produces redundant/incoherent groupings in the imbalanced data.  In contrast, our model is able to discover digit identities in the imbalanced data.}  
    \label{fig:infogan}
\end{figure*}

InfoGAN \cite{chen-nips16} in particular, learns an unsupervised disentangled representation by maximizing the mutual information between the discrete or continuous latent variables and the corresponding generated samples. For discrete latent factors (e.g., digit identities), it assumes that they are uniformly distributed in the data, and approximates them accordingly using a \emph{fixed uniform} categorical distribution. Although this assumption holds true for many benchmark datasets (e.g., MNIST~\cite{lecun-mnist1998}), real-word data often follows a long-tailed distribution and rarely exhibits perfect balance between the categories.  Indeed, applying InfoGAN on imbalanced data can result in incoherent groupings, since it is forced to discover potentially non-existent factors that are uniformly distributed in the data; see Fig.~\ref{fig:infogan}. 

In this work, we augment InfoGAN to discover disentangled categorical representations from \emph{imbalanced} data.  Our model, Elastic-InfoGAN, makes two improvements to InfoGAN which are simple and intuitive.  First, we remodel the way the latent distribution is used to fetch the latent variables; we lift the assumption of any knowledge about the underlying class distribution, where instead of deciding and fixing them beforehand, we treat the class probabilities as \emph{learnable parameters} of the optimization process. To enable the flow of gradients back to the class probabilities, we employ the Gumbel-Softmax distribution \cite{jang-iclr2017,maddison-iclr2017}, which acts as a proxy for the categorical distribution, generating \emph{differentiable} samples having properties similar to that of categorical samples.   
Our second improvement stems from an observation of a failure case of InfoGAN (Fig.~\ref{fig:infogan} center); we see that the model has trouble generating consistent images from the same category for a latent dimension (e.g., rows 1, 2, 4). This indicates that there are other low-level factors (e.g., rotation, thickness) which the model focuses on while categorizing the images. Although there are multiple meaningful ways to partition unlabeled data---e.g., with digits, one partitioning could be based on identity, whereas another could be based on stroke width---we aim to discover the partitioning that groups objects according to a high-level factor like identity while being invariant to low-level ``nuisance'' factors like lighting, pose, and scale changes. To this end, we take inspiration from self-supervised contrastive representation learning literature {\cite{bachman-neurips2019, henaff-arxiv2019,chen-arxiv2020}} to learn representations focusing on \emph{object identity}. Specifically, we enforce (i) similar representations for positive pairs (e.g., an image and its mirror-flipped version), and (ii) dissimilar representations for negative pairs (e.g., two different images). As a result, the discovered latent factors align more closely with object identity, and less with other factors. Such partitionings focusing on object identity are more likely to be useful for downstream visual recognition applications; e.g. (i) semi-supervised object recognition~\cite{radford-iclr16,noroozi-eccv16} or image retrieval using object-identity based image features; (ii) performing data augmentation to remove class-imbalance using synthetic images.
Importantly, Elastic-InfoGAN retains InfoGAN's ability to jointly model both continuous and discrete factors in either balanced or imbalanced data scenarios. To our knowledge, our work is the first to tackle the problem of disentangled representation learning in the scenario of imbalanced data, \emph{without} the knowledge of ground-truth class distribution (Fig.~\ref{fig:infogan} right). 
We show qualitatively and quantitatively our superiority in terms of the ability to disentangle object identity as a factor of variation, in comparison to relevant baselines. And in order to discover object identity as a factor, our results also provide interesting observations regarding the \emph{ideal} distribution for the latent variables.


\section{Related Work}

\textbf{Disentangled representation learning~}  has a vast literature \cite{hinton-icann2011,bengio-tpami2013,yan-eccv16,chen-nips16,mathieu-nips2016,tran-cvpr2017,denton-nips17,hu-cvpr18,singh-cvpr2019}. In particular, InfoGAN \cite{chen-nips16} learns disentanglement without supervision by maximizing the mutual information between the latent codes and generated images, and has shown promising results for \emph{class-balanced} datasets like MNIST \cite{lecun-mnist1998}, CelebA \cite{liu-iccv15}, and SVHN \cite{netzer-nipsw2011}. JointVAE \cite{dupont-neurips2018} extends beta-VAE \cite{higgins-iclr17} by jointly modeling both continuous and discrete factors, using Gumbel-Softmax sampling. However, both InfoGAN and JointVAE assume uniformly distributed data, and hence fail to be equally effective in imbalanced data, evident by Fig.~\ref{fig:infogan} and our experiments.
Our work proposes improvements to InfoGAN to enable it to discover meaningful latent factors in \emph{imbalanced} data.


\textbf{Learning from imbalanced data~} Real world data have a long-tailed distribution \cite{guo-eccv2016,van-cvpr2018}, which can impede learning, since the model can get biased towards the dominant categories. Re-sampling \cite{chawla-jair2002,he-ijcnn2008, shen-eccv2016,buda-nn2018,zou-eccv18} and class re-weighting techniques \cite{ting-icml2000,huang-cvpr2016,dong-iccv2017,mahajan-eccv2018} can alleviate this issue for the \emph{supervised} setting, in which the class distributions are known a priori.  There are also unsupervised clustering methods that deal with imbalanced data in unknown class distributions (e.g., \cite{nguwi-esa2010,you-eccv2018}).  Our model works in the same \emph{unsupervised} setting; however, unlike these methods, we propose a \emph{generative} model method that learns to disentangle latent categorical factors in imbalanced data.

\textbf{Data augmentation for unsupervised image grouping~} Unsupervised deep clustering methods~\cite{dizaji-iccv17,yang-cvpr16, shah-arxiv18, xie-icml16} try to group unlabeled instances that belong to the same object category. Some works \cite{hui-icml2013,dosovitskiy-tpami2015,hu-icml2017,ji-iccv2019,bachman-neurips2019,chen-arxiv2020} use data augmentation for image transformation invariant clustering or representation learning. The main idea is to maximize the mutual information or similarity between the features of an image and its corresponding transformed image. Some approaches also try to make different images more dissimilar~\cite{bachman-neurips2019,chen-arxiv2020}.  However, unlike our approach, these methods typically do not target imbalanced data and do not perform generative modeling.

\textbf{Differentiable approximation of categorical variables~} Existence of discrete random variables within a computation graph introduces non-differentiability. Consequently, recent works have introduced a reparameterization trick using Gumbel-Softmax, enabling differentiable sampling of variables which approximate the categorical ones \cite{jang-iclr2017,maddison-iclr2017}. This has been used in neural architecture search \cite{wu-cvpr2019,xie-iclr2019,wu-arxiv2019}, where it makes the process of choosing blocks (e.g. out of $k$ distinct options) for a layer differentiable, enabling the overall search process possible through gradient-based optimization methods. Recently, it has also found use-cases in approximating discrete segmentation masks for scene generation~\cite{azadi-arxiv2019}. Our work uses Gumbel-Softmax reparameterization for a different application, where we seek to learn a better multinomial distribution for discrete factors (e.g. object identity) in class-imbalanced data.



\begin{figure*}[t]
    \centering
    \includegraphics[width=1\textwidth]{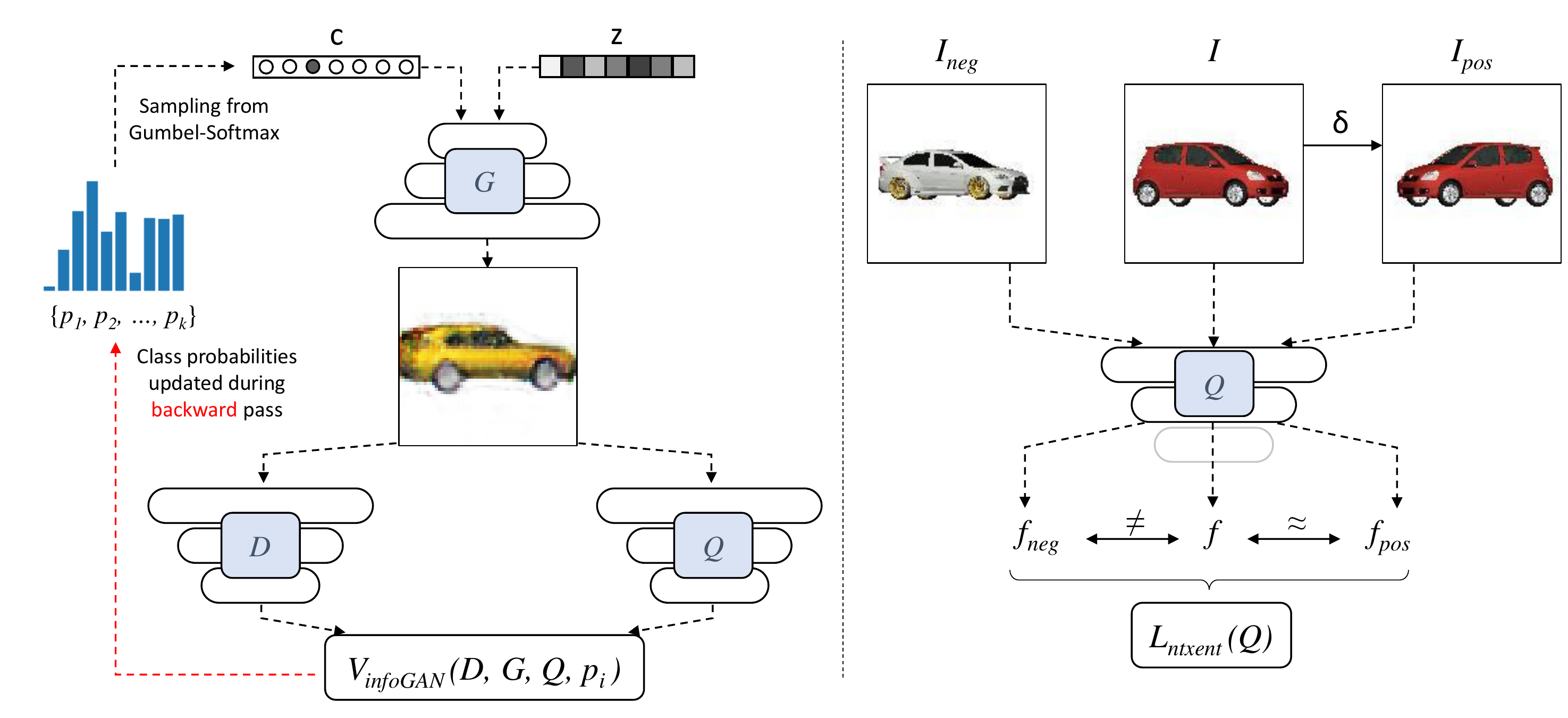}
    \vspace{-5pt}
    \caption{(\textbf{Left}) Our model takes a sampled categorical code from a Gumbel-Softmax distribution and a noise vector to generate fake samples. The use of differentiable latent variables from the Gumbel-Softmax enables gradients to flow back to the class probabilities to update them.  (\textbf{Right}) Apart from the original InfoGAN~\cite{chen-nips16} losses, we have an additional loss for contrastive learning of representations. We transform a real image $I$ using commonly used data augmentations $\delta$ (e.g. mirror flipping, random crop) to create a positive pair $\{I, I_{pos}\}$, and enforce similarity in their respective features extracted using $Q$. The same real image is also paired up with an arbitrary image from the same batch to create a negative pair $\{I, I_{neg}\}$, whose feature representations are made dissimilar.}
    \label{fig:main}
\end{figure*}

\section{Approach}

Let $\mathcal{X} = \{x_1, x_2,\dots,x_N\}$ be a dataset of $N$ unlabeled images from $k$ different classes. No knowledge about the nature of class imbalance is known beforehand. 
Our goal is to learn a generative model $G$ which can learn to disentangle object category from other aspects (e.g., digits in MNIST \cite{lecun-mnist1998}, face identity in YouTube-Faces \cite{wolf-cvpr2011}) in imbalanced data, by approximating the appropriate latent distribution. 
In the following, we first briefly discuss InfoGAN \cite{chen-nips16}, which addressed this problem for the balanced setting. We then explain how it can be improved to handle imbalanced data.



\subsection{Background: InfoGAN}
Learning disentangled representations using the GAN \cite{goodfellow-nips2014} framework was introduced in InfoGAN \cite{chen-nips16}. The intuition is for the generated samples to retain information about latent variables, and consequently for latent variables to gain control over certain aspects of the generated image. In this way, different types of latent variables (e.g., discrete categorical vs.~continuous) can control properties like discrete (e.g., digit identity) or continuous (e.g., digit rotation) variations in the generated images. 

Formally, InfoGAN does this by maximizing the mutual information between the latent code $c$ and the generated samples $G(z, c)$, where $z \sim P_{noise}(z)$ and $G$ is the generator network. The mutual information $I(c, G(c,z))$ can then be used as a regularizer in the standard GAN training objective. Computing $I(c, G(c,z))$ however, requires $P(c|x)$, which is intractable and hard to compute. The authors circumvent this by using a lower bound of $I(c, G(c,z))$, which can approximate $P(c|x)$ via a neural network based auxiliary distribution $Q(c|x)$. The training objective hence becomes:
\begin{align}
\vspace{-5pt}
   \min\limits_{G, Q}\max\limits_{D}{V}_{InfoGAN}(D, G, Q) = V_{GAN}(D, G) - \lambda_1 L_{1}(G, Q),\label{eq:infogan}
\end{align}   
where $L_{1}(G, Q) = E_{c \sim P(c), x \sim G(z, c)}[\log Q(c|x)] + H(c)$, $D$ is the discriminator network, and $H(c)$ is the entropy of the latent code distribution.
Training with this objective results in latent codes $c$ having control over the different factors of variation in the generated images $G(z, c)$. To model discrete variations in the data, InfoGAN employs non-differentiable samples from a uniform categorical distribution with fixed class probabilities; i.e., $ c \sim Cat(K = k, p = 1/k)$ where $k$ is the number of discrete categories to be discovered. 
%

\vspace{-1pt}
\subsection{Disentangling object identity in imbalanced data}\label{sec:approach}
\vspace{-1pt}

As shown in Fig.~\ref{fig:infogan}, applying InfoGAN to an imbalanced dataset results in suboptimal disentanglement, since the uniform prior assumption does not match the actual ground-truth data distribution of the discrete factor (e.g.,~digit identity). 
To address this, we propose two improvements to InfoGAN. The first is to enable \emph{learning} of the latent distribution's parameters (class probabilities), which requires gradients to be backpropagated through latent code samples $c$, and the second involves constrastive learning of representations, so that the discovered factor aligns closely with \emph{object identity}. 

\textbf{Learning the prior distribution~} To learn the prior distribution, we replace the fixed categorical distribution in InfoGAN with the Gumbel-Softmax distribution \cite{jang-iclr2017,maddison-iclr2017}, which enables sampling of differentiable samples. The continuous Gumbel-Softmax distribution can be smoothly annealed into a categorical distribution. Specifically, if $p_1, p_2 ..., p_k$ are the class probabilities, then sampling of a $k$-dimensional vector $c$ can be done in a differentiable way:
\begin{equation}
c_i = \frac{\text{exp}((\log(p_i)+g_i)/\tau)}{\sum_{j=1}^k \text{exp}((\log(p_j)+g_j)/\tau)} \qquad \text{for } i=1, ..., k.
\end{equation}
Here $g_i, g_j$ are samples drawn from $Gumbel(0, 1)$, and $\tau$ (softmax temperature) controls the degree to which samples from Gumbel-Softmax resemble the categorical distribution. Low values of $\tau$ make the samples possess properties close to that of a one-hot sample.   

In theory, InfoGAN's behavior in the class balanced setting (Fig.~\ref{fig:infogan} left) can be replicated in the imbalanced case (where grouping becomes incoherent, Fig.~\ref{fig:infogan} center), by simply replacing the fixed uniform categorical distribution with Gumbel-Softmax with \emph{learnable} class probabilities $p_i$'s; i.e. gradients can flow back to update the class probabilities (which are uniformly initialized) to match the true class imbalance. And once the true imbalance gets reflected in the class probabilities, the \emph{possibility} of proper categorical disentanglement (Fig.~\ref{fig:infogan} right) becomes feasible.

\begin{wrapfigure}[5]{r}{0.3\textwidth}
	\centering
	\vspace{-10pt}
	\includegraphics[width=0.25\textwidth]{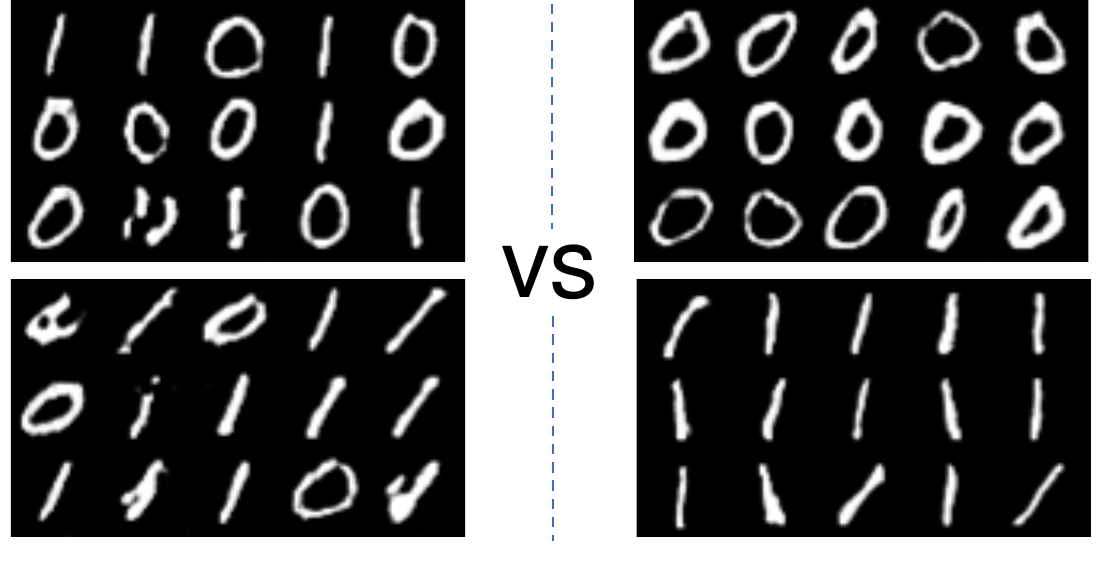}
\end{wrapfigure}
Empirically, however, this ideal behavior is not observed in a consistent manner. As shown in the figure on the right, unsupervised grouping can focus on non-categorical attributes such as rotation of the digit (left). Although this is one valid way to group unlabeled data, our goal is to have groupings that correspond to \emph{class identity} (right). This would enable useful applications for downstream tasks such as semi-supervised image classification, removing class-imbalance with generated data, and identity-based image retrieval.





\textbf{Learning object identities~} Based on Eq.~\ref{eq:infogan}, the factor of variation discovered by the latent vector $c$ will depend on the factor that $Q$ focuses on while making the classification decision (whether images are assigned different classes based on e.g., pose, illumination, identity, etc.). So, to enable the discovery of object identity as the factor, we enforce $Q$ to learn representations using a contrastive loss~\cite{hadsell-cvpr06}. The idea is to create positive pairs (e.g., a car and it's mirror flipped version) and negative pairs (e.g., a red hatchback and a white sedan) based on object identity, and have $Q$ produce similar and dissimilar representations for them respectively.

Since we do not have any category labels, we treat each instance (image) as its own class. Intuitively, and as validated by prior work~\cite{bachman-neurips2019,chen-arxiv2020}, forcing a model to predict the same feature for two different views (augmentations) of the same image leads to the model learning a representation focusing on high-level factors like object identity.  
Formally, for a sampled batch of $N$ real images, we construct their augmented version, by applying \emph{identity preserving} transformations ($\delta$) to each image, resulting in a total of $2N$ images. For each image $I_i$ in the batch, we define the corresponding transformed image as $I_{pos}$, and all other $2(N-1)$ images as $I_{neg}$. We define the following loss for each image $I_i$ in the batch, where $\{I_i, I_{pos}\}$ and $\{I_i, I_{neg}\}$ act as positive and negative pairs, respectively:   
\begin{equation}
\ell_{i}=-\log \frac{\exp \left(\operatorname{sim}(f_{i}, f_{j}) / \tau{^{'}}\right)}{\sum_{k=1}^{2N} \mathbf{1}_{[k \neq i]} \exp \left(\operatorname{sim}(f_{i}, f_{k}) / \tau{^{'}}\right)}
\label{eq:ntxent}
\end{equation}
where $j$ indexes the positive pair, $f$ represents the feature extracted using $Q$ (we use the penultimate layer), $\tau^{'}$ is a softmax temperature, and sim(.) refers to cosine similarity.
Note that since any two (unlabeled) images in a batch, except $I_i$ and $I_{pos}$, are treated as negative pairs, there would be cases where the sampled pair will be a false negative (i.e., images belonging to the same category). 
However, the fraction of false negatives remains considerably low except for highly-skewed data scenarios (we provide analysis in the supplementary), and even then we take the penultimate layer's features for computing similarity/dissimilarity which provides some robustness to such errors as the final layer can still put two false negatives to be similar to each other. These aspects make the above approximation of sampling negative pairs practically applicable in our setting. 
We denote the overall loss as $L_{ntxent} = \sum_{i=1}^{N} \ell_{i}$ (normalized temperature-scaled cross entropy loss~\cite{chen-arxiv2020}). Our training objective hence becomes:
\begin{equation}\label{eqn:final}
    \min\limits_{G, Q, p_{i}}\max\limits_{D} L_{final} = {V}_{InfoGAN}(D, G, Q, p_{i}) + \lambda_2 L_{ntxent}(Q).
\end{equation}
${V}_{InfoGAN}$ plays the role of generating realistic images and associating the latent variables to correspond to \emph{some} factor of variation in the data, while the addition of $L_{ntxent}$ will push the discovered factor of variation to be close to object identity. 
The latent codes sampled from Gumbel-softmax, generated fake images, and losses operating on fake images are all functions of class probabilities $p_i$'s too. Thus, during the minimization phase of $L_{final}$, the gradients are used to optimize the class probabilities along with $G$ and $Q$ in the backward pass. Overall, we leverage previous ideas used in orthogonal areas: Gumbel-Softmax was introduced for differentiable sampling of one-hot like variables, and data augmentations have been used for various visual recognition tasks. Our framework integrates them in a coherent way to address the new problem of generative modeling of latent object identity factor in class-imbalanced data.

\begin{figure*}[t!]
    \centering
    \includegraphics[width=0.8\textwidth]{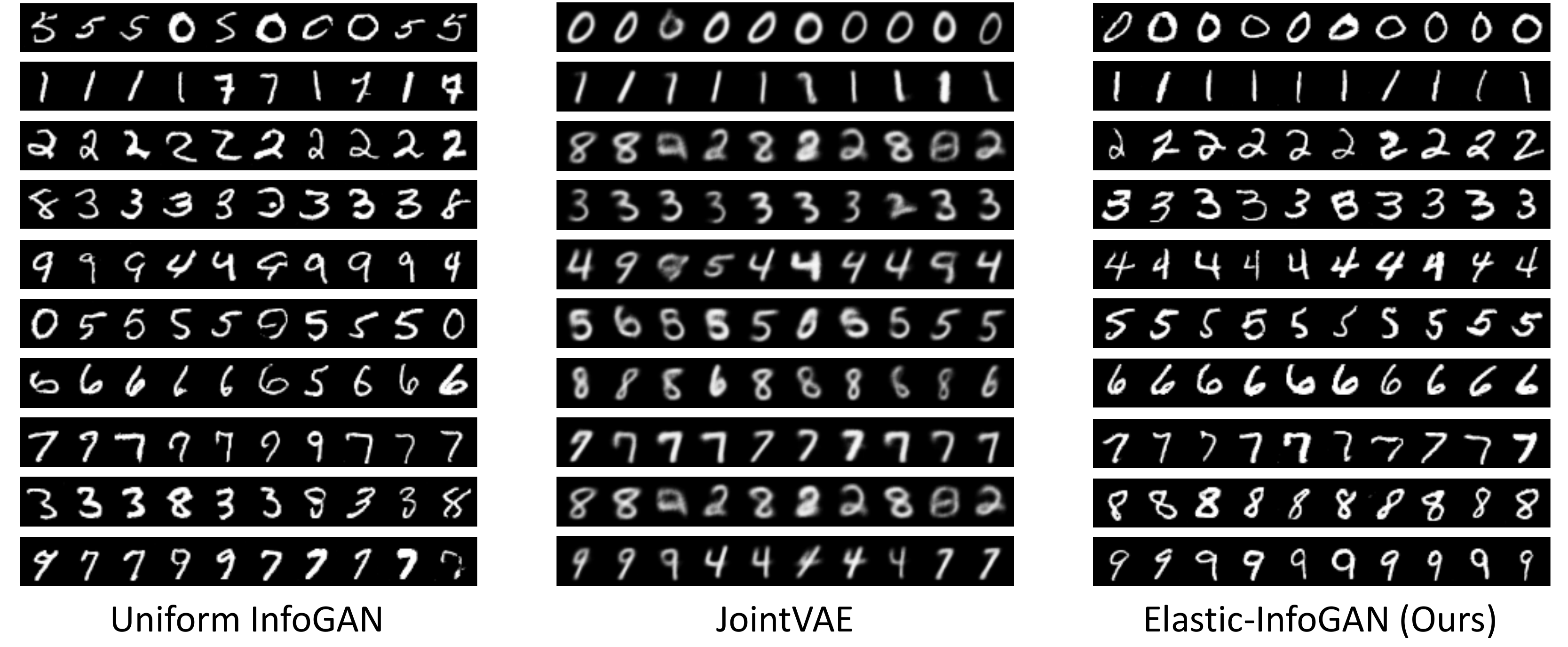}
    \caption{Representative generations on a random imbalanced MNIST split. Each row corresponds to a learned latent variable. Elastic-InfoGAN generates inconsistent images in only the 4th row (8 with 3's), whereas Uniform InfoGAN and JointVAE do so in many rows (e.g. rows 1, 9 and rows 3, 8 respectively). 
    }
    \label{fig:mnist}
\end{figure*}

\section{Experiments}

In this section, we perform quantitative and qualitative analyses to demonstrate the advantage of our model in discovering categorical disentanglement for imbalanced datasets.

\textbf{Datasets~} 
\textbf{(1) MNIST} \cite{lecun-mnist1998} is by default a balanced dataset with 70k images, with a similar number of training samples for each of 10 classes. We artificially introduce imbalance over 50 random splits, and report the results averaged over them.
\textbf{(2) 3D Cars}~\cite{fidler-neurips12} and \textbf{(3) 3D Chairs}~\cite{aubry-cvpr14} consist of synthetic objects rendered with varying identity and poses. We choose 10 random object categories for both 3D cars (out of 183 total categories) and 3D chairs (out of 1396 total categories), where each category contains 96 and 62 images for cars and chairs, respectively. Within the chosen 10 categories, we introduce imbalance over 5 random splits. The whole process is repeated 5 times (choosing a different set of 10 categories randomly each time) so as to test the generalizability of the approach. \textbf{(4) ShapeNet} is a dataset of 3D models of diverse object categories, whose 2D renderings can be obtained in different pose/viewpoints. We choose 5 categories (synsets) from ShapeNetCore - cars, airplanes, bowl, can, rifle - which are more diverse in object shape/appearance compared to 3D Cars/Chairs. Each category has a large number of instances (different car models within the \emph{cars} synset). We select different number of instances for each of the categories to introduce imbalance, and generate 30 renderings \footnote{we use - \texttt{https://github.com/panmari/stanford-shapenet-renderer}} for each instance in different viewpoints. \textbf{(5) YouTube-Faces} \cite{wolf-cvpr2011} is a real world imbalanced dataset with varying number of training samples (frames) for 40 face identity classes (as used in \cite{shah-arxiv18}). The smallest/largest class has 53/695 images, with a total of 10,066 tightly-cropped face images.  The results are reported over the average of 5 runs over the same imbalanced dataset. (The imbalance statistics for all datasets are in the supplementary).

\begin{figure*}[t!]
    \centering
    \includegraphics[width=1\textwidth]{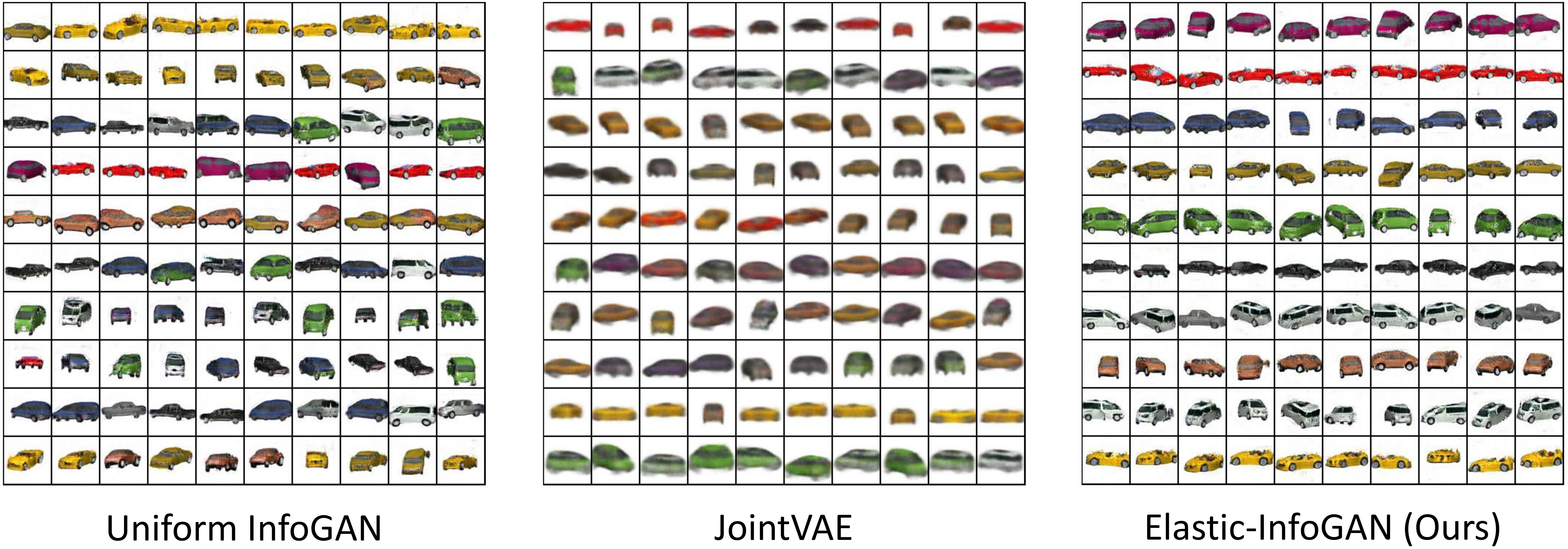}
    \caption{Image generations on a random imbalanced 3D Cars split, on a randomly chosen sets of categories. Each row corresponds to a learned latent variable. The images are much more consistent corresponding to a latent variable for Elastic-InfoGAN,  compared with Uniform InfoGAN. JointVAE struggles in the aspects of realism as well as consistency among the generations for a latent code.}
    \label{fig:3d_cars}
\end{figure*}

\textbf{Baselines and evaluation metrics~} We design different baselines to show the importance of different components of our approach. \textbf{(i) \emph{Uniform InfoGAN}} \cite{chen-nips16}: This is the original InfoGAN with fixed and uniform categorical distribution. \textbf{(ii) \emph{Ground-truth InfoGAN}}: This is InfoGAN with a fixed, but imbalanced categorical distribution where the class probabilities reflect the ground-truth class imbalance. \textbf{(iii) \emph{Ground-truth InfoGAN + $L_{ntxent}$}}: Similar to the previous baseline but with the contrastive loss (Eq.~\ref{eq:ntxent}). \textbf{(iv) \emph{Gumbel-softmax}}: Similar to InfoGAN, but this baseline does not have a fixed prior for the latent variables.  Instead, the priors are learned using the Gumbel-softmax technique \cite{jang-iclr2017}. \textbf{(v) \emph{Gumbel-softmax + pos-$L_{ntxent}$}}: This is the version where apart from having a learnable prior, we also apply a part of $L_{ntxent}$, where we enforce the positive pairs to have similar latent prediction $Q(c|x)$ but do not use negative pairs. \textbf{(vi) \emph{Elastic-InfoGAN}}: This is our final model in which we use the complete form of $L_{ntxent}$. \textbf{(vii) \emph{JointVAE}} \cite{dupont-neurips2018}: We also include this VAE based baseline, which performs joint modeling of disentangled discrete and continuous factors. The objective function uses two KL-divergence loss terms to enforce the inferred discrete and continuous latent variables to follow their respective prior distributions (e.g., uniform categorical and standard normal respectively). We tune the weights for both of these loss functions separately, and report the best configuration's results (see supplementary for more details).

The standard metrics for evaluating disentanglement require access to the ground-truth latent factors ~\cite{eastwood-iclr18,kim-icml18,higgins-iclr17}, which is rarely the case with real-world datasets. Furthermore, our evaluation should specifically capture the ability to disentangle \emph{class-specific} information from other factors in an imbalanced dataset. Since the aforementioned metrics don't capture this property, we propose to use the following metrics:  
\textbf{(a) \emph{Average Entropy (ENT)}:} Evaluates two properties: (i) whether the images generated for a given categorical code belong to the same ground-truth class i.e., whether the ground-truth class histogram for images generated for each categorical code has low entropy; (ii) whether each ground-truth class is associated with a single unique categorical code. We generate 1000 images for each of the $k$ latent categorical codes, compute class histograms using a pre-trained classifier\footnote{We train the classifier by creating a 80/20 train/val split on a per class basis. Classification accuracies: (i) MNIST: 98\%, (ii) 3D Cars: 99\%, (iii) 3D Chairs: 97\%, (iv) ShapeNet: 95\%, (v) YouTube-Faces: 96\%. See supplementary for details.} to get a $k \times k$ matrix (where rows index latent categories and columns index ground-truth categories). We report the average entropy across the rows (tests (i)) and columns (tests (ii)). \textbf{(b) \emph{Normalized Mutual Information (NMI)}} \cite{xu-sigir03}: We treat our latent category assignments of the fake images (we generate 1000 fake images for each categorical code) as one clustering, and the category assignments of the fake images by the pre-trained classifier as another clustering.  NMI measures the correlation between the two clusterings. The value of NMI will vary between 0 to 1; higher the NMI, stronger the correlation.

\begin{figure*}[t!]
    \centering
    \includegraphics[width=1\textwidth]{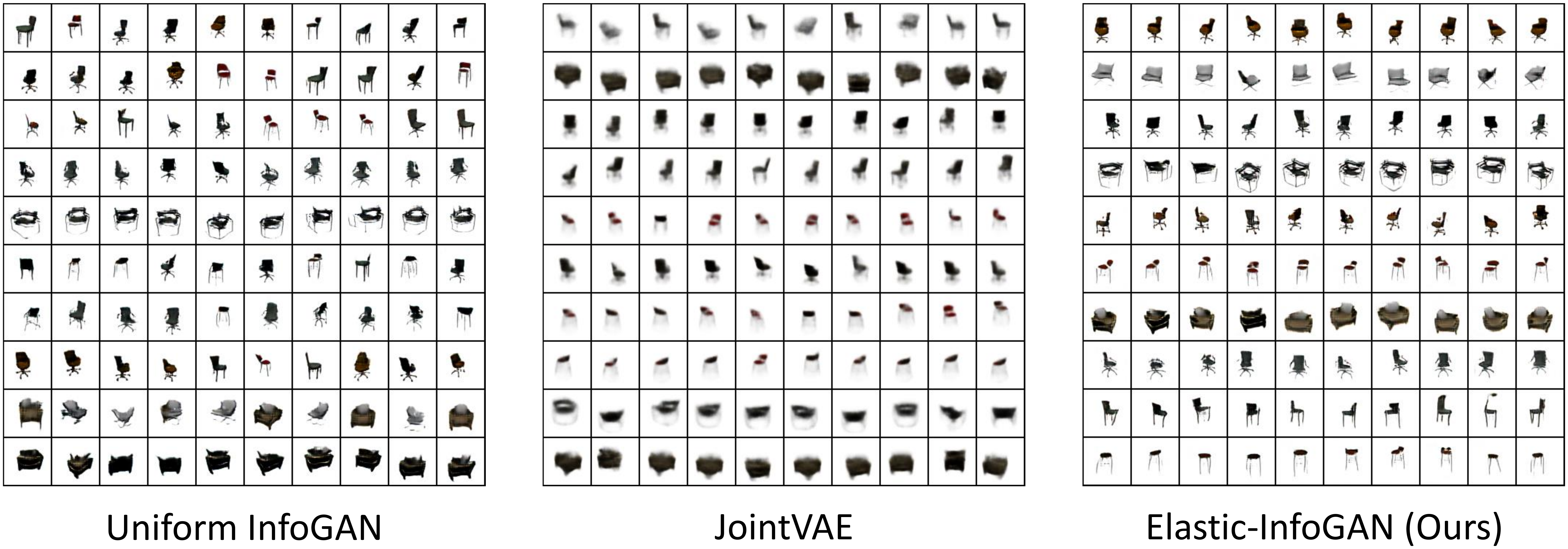}
    \caption{Image generations on a random imbalanced 3D Chairs split, on a randomly chosen sets of categories. Each row corresponds to a learned latent variable. Similar to the results on existing datasets, images are much more consistent corresponding to a latent variable for Elastic-InfoGAN, compared with Uniform-InfoGAN or JointVAE. Due to lack of details in the results for JointVAE, it is hard to figure out if some categories repeat in multiple rows.}
    \label{fig:3d_chairs}
\end{figure*}

\textbf{Implementation details~} Transformations ($\delta$) used: (i) MNIST: Rotation ($\pm 10$ deg) + Zoom ($\pm 0.1 \times$); for (ii) 3D Cars, (iii) 3D Chairs, and (iv) ShapeNet: Rotation ($\pm 10$ deg) + Random horizontal flip + Random crop (preserving 95\% of image); (v) YouTube-Faces: Random horizontal flip + Random crop (scale image by $1.1\times$ and crop $64 \times 64$ patch) + Gamma contrast (gamma $\sim U(0.3, 4.0)$). Refer to supplementary for more implementation details.

\begin{table}[]
\centering
\scriptsize
\begin{tabular}{l|c|c|c|c|c|c|c|c|c|c}

                                    & \multicolumn{2}{c|}{MNIST} & \multicolumn{2}{c|}{YTF} & \multicolumn{2}{c|}{3D-cars} & \multicolumn{2}{c|}{3D-chairs} & \multicolumn{2}{c}{ShapeNet} \\ \hline
                                    & NMI          & ENT         & NMI         & ENT        & NMI           & ENT          & NMI            & ENT       & NMI            & ENT    \\ \hline
JointVAE \cite{dupont-neurips2018}      & 0.704       & 0.661      & 0.485      & 1.554     & 0.458        & 1.024       & 0.480         & 1.817   & 0.189  & 1.101    \\ 
Uniform-InfoGAN \cite{chen-nips16}                    & 0.777       & 0.457      & 0.666      & 1.031     & 0.499        & 1.108       & 0.253         & 1.663  & 0.638 & 0.531      \\ 
Gumbel-Softmax                      & 0.836       & 0.326      & 0.760      & 0.757     & 0.400        & 1.318       & 0.236         & 1.696    & 0.603 & 0.619     \\ 
Gumbel-Softmax + pos-$L_{ntxent}$        &     0.878   &   0.235    & 0.765      & 0.719     &  0.582       &  0.919      &    0.454      & 1.209   & 0.724 & 0.397    \\ 
Elastic-InfoGAN (Ours)     & \textbf{0.889}       & \textbf{0.213}      & \textbf{0.792}      & \textbf{0.636}     & \textbf{0.850}        & \textbf{0.303}       & \textbf{0.650}         & \textbf{0.765}  & \textbf{0.790} &    \textbf{0.297}   \\ \hline
Ground-truth InfoGAN                & 0.783       & 0.412      & 0.694      & 0.961     & 0.451        & 1.191       & 0.174         & 1.837    & 0.549 & 0.673    \\ 
Ground-truth InfoGAN + $L_{ntxent}$  & 0.801       & 0.369      & 0.742      & 0.767     & 0.784        & 0.437       & 0.592         & 0.885      & 0.531 &  0.716 \\ \hline

\end{tabular}
\vspace{3pt}
\caption{Distentanglement quality measured by NMI ($\uparrow$) and ENT ($\downarrow$). The first five methods have no knowledge of the ground-truth distribution, while the last two methods do. We see that incorporating contrastive loss within a baseline (either Gumbel-Softmax or Ground-truth InfoGAN) helps the model better learn the disentangled representations. Overall, Elastic-InfoGAN demonstrates the ability to better disentangle object identity from other factors compared to baselines. (See supplementary for error bars.)} 
\vspace{-8pt}
\label{table:metrics}
\end{table}

\subsection{Quantitative evaluation}
 
\textbf{Comparisons to baselines assuming uniform prior~} As explained in previous sections, baselines using this prior for the latent categorical distribution would (in theory) have difficulty in disentangling object identity as a separate factor in class imbalanced data. We observe this behavior empirically too (see Table~\ref{table:metrics}); Elastic-InfoGAN, which learns the latent categorical distribution, obtains significant boosts of 0.113 and 0.127 in NMI, and -0.244 and -0.395 in ENT compared to the Uniform InfoGAN baseline for MNIST and YouTube-Faces, respectively.  The boost is even more significant when compared to JointVAE: 0.209, 0.345 in NMI, and -0.4877, -0.1.081 in ENT for MNIST and YouTube-Faces, respectively. 
Similar results can be observed for Cars, Chairs, and ShapeNet, where our method gets a boost of 0.351, 0.397, 0.152 respectively in NMI, and -0.805, -0.897, -0.234 respectively in ENT, compared to Uniform InfoGAN. 

\textbf{Effect of using contrastive loss~} From Table~\ref{table:metrics}, we see that the performance of baselines learning the prior distribution follow a particular order across all the datasets. Having an auxiliary constraint to enforce positive pairs to have similar representations improves performance (Gumbel-Softmax vs. Gumbel-Softmax + pos-$L_{ntxent}$), and additionally constraining the negative pairs to have dissimilar representations results in further gains (Gumbel-Softmax + pos-$L_{ntxent}$ vs. Elastic-InfoGAN). This trend indicates that the absence of appropriate auxiliary constraints results in the model having no signal to be invariant to undesirable factors. 
For example, one of the ways in which different `ones' in MNIST vary is rotation, which can be used as a factor (as opposed to object identity) to group data in imbalanced cases (recall the different ways to group from Sec.~\ref{sec:approach}). Similarly in Cars, if the number of different poses actually match the number of discrete categories, pose could emerge as a factor instead of object identity. These are the potential scenarios where Gumbel-Softmax/Ground-truth InfoGAN will perform more poorly than Elastic-InfoGAN (see supplementary for more results on this). Note that the transformations ($\delta$) used in $L_{ntxent}$ are not supposed to capture all the intra-class variations themselves; their role is to help the model ($Q$) focus more on object identity while predicting the latent category, by ruling out the variations in $\delta$ as a way to group. 

%


\begin{figure*}[t!]
    \centering
    \includegraphics[width=1\textwidth]{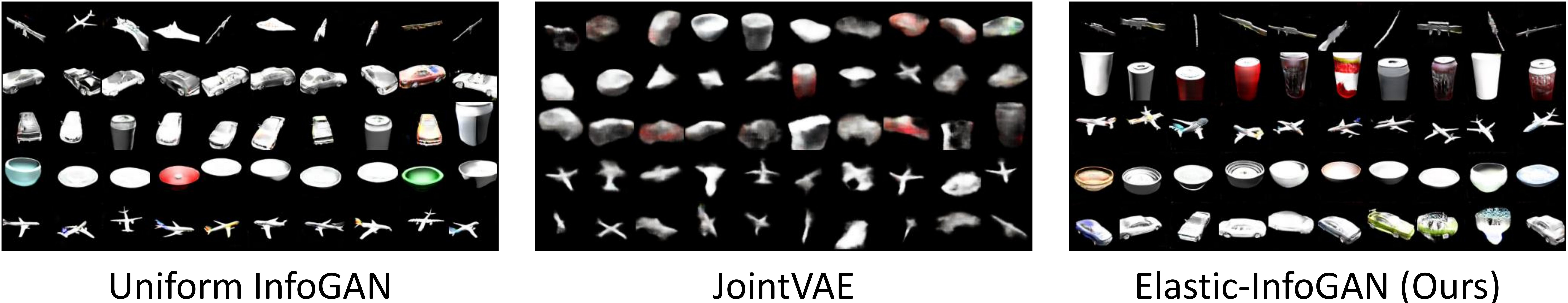}
    \caption{Image generations on ShapeNet: different categorical latent codes capture object identity (e.g., planes/cars) much more consistently in Elastic-InfoGAN than Uniform-InfoGAN, which mixes up generations (e.g., cars with cans in row 3). JointVAE shares this issue, in addition to poor quality of generations.}
    \label{fig:shapenet}
\end{figure*}

Interestingly, using a fixed ground-truth prior (Ground-truth InfoGAN) does not always result in better disentanglement than learning the prior (Gumbel-softmax). This requires further investigation, but we hypothesis an explanation based on the idea presented in~\cite{cui-2019}. The \emph{effective number of samples} in a category will not necessarily be the same as the number of instances in that category; e.g., two 0 digits that are almost equivalent to each other would result in an effective sample size closer to one instance rather than two instances. 
It is therefore possible that the distribution for the effective samples might not exactly match the ground-truth imbalanced distribution of instances, and hence using the ground-truth distribution for the latent space might result in sub-optimal disentanglement. 




\begin{figure*}[t!]
    \centering
    \includegraphics[width=1\textwidth]{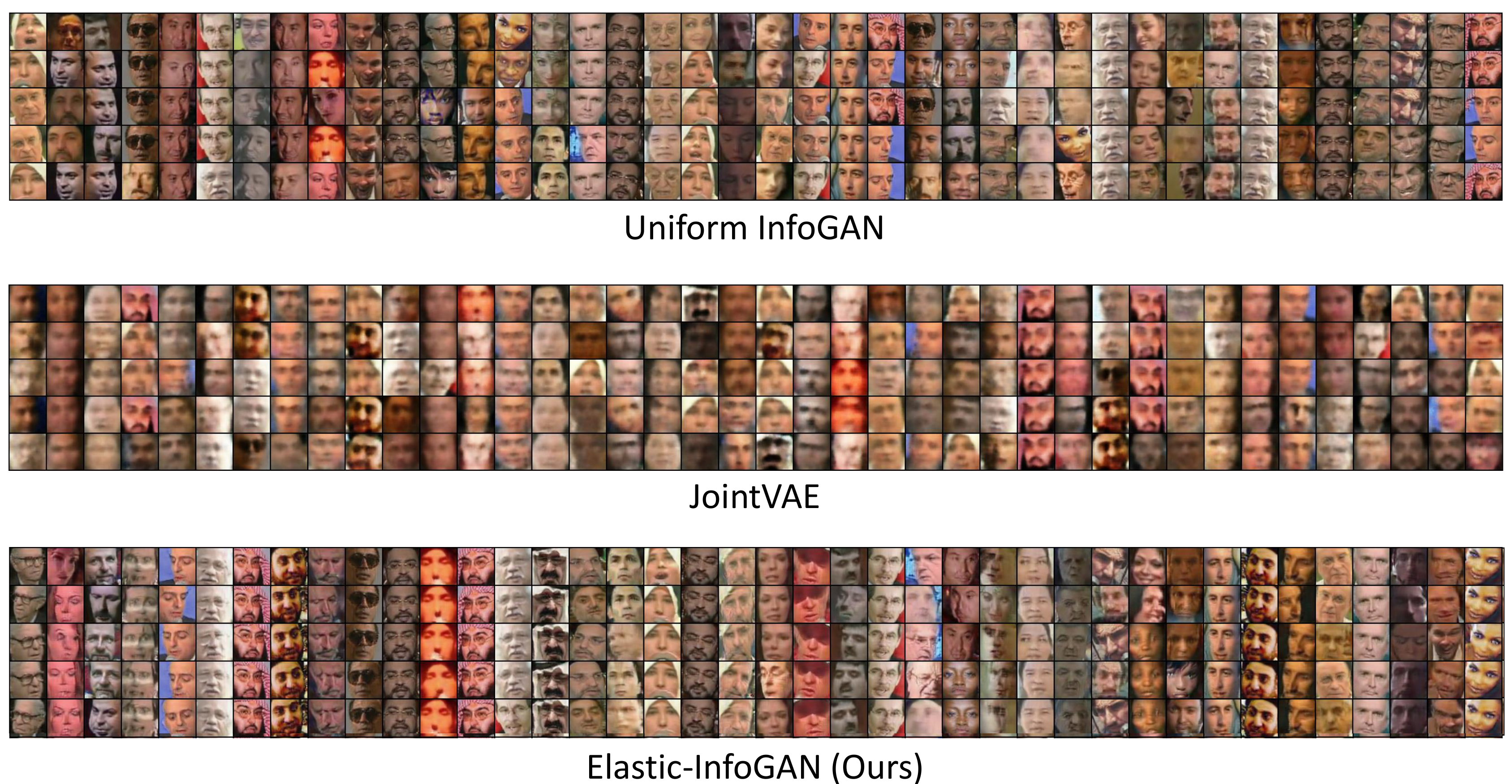}
    \caption{Image generations on YouTube-Faces. Each column corresponds to a latent variable. Although there are a few redundant latent variables (e.g., 7th and 13th columns) in Elastic-InfoGAN, it generates images belonging to the same person more consistently compared to Uniform-InfoGAN and JointVAE, which tend to mix up face identities a lot more frequently. 
    }
    \label{fig:face}
\end{figure*}

\subsection{Qualitative evaluation}

We next qualitatively evaluate our method's disentanglement ability. Figs.~\ref{fig:mnist}-\ref{fig:face} show results for MNIST, Cars, Chairs, ShapeNet, and YouTube-Faces.  Overall, Elastic-InfoGAN generates more consistent images for each latent code compared to Uniform InfoGAN and JointVAE. For example, in Fig.~\ref{fig:mnist}, our model only generates inconsistent images in the 4th row (mixing up 8 with 3's) whereas the baselines generate inconsistent images in several rows. In Fig.~\ref{fig:3d_cars} and Fig.~\ref{fig:3d_chairs}, we see that for a given latent variable, our model can consistently generate images from the same object category, in different pose/viewpoints. For Chairs, there are some cases in which the images generated in different rows look similar for Elastic-InfoGAN, but this still happens much less frequently than Uniform InfoGAN and JointVAE. Fig.~\ref{fig:shapenet} further demonstrates the ability of Elastic-InfoGAN to become invariant to other forms of continuous factors (e.g. object pose), with different categorical codes accurately representing different high-level object categories; e.g,. rifles vs cars, in a better way than the baseline methods. Similarly, in Fig.~\ref{fig:face}, our model generates faces of the same person corresponding to a latent variable more consistently than the baselines. Both Uniform InfoGAN and JointVAE, on the other hand, more often mix up identities within the same categorical code because they incorrectly assume a prior uniform distribution. 

\begin{wrapfigure}[7]{r}{0.4\textwidth}
	\centering
	\vspace{-11pt}
	\includegraphics[width=0.4\textwidth]{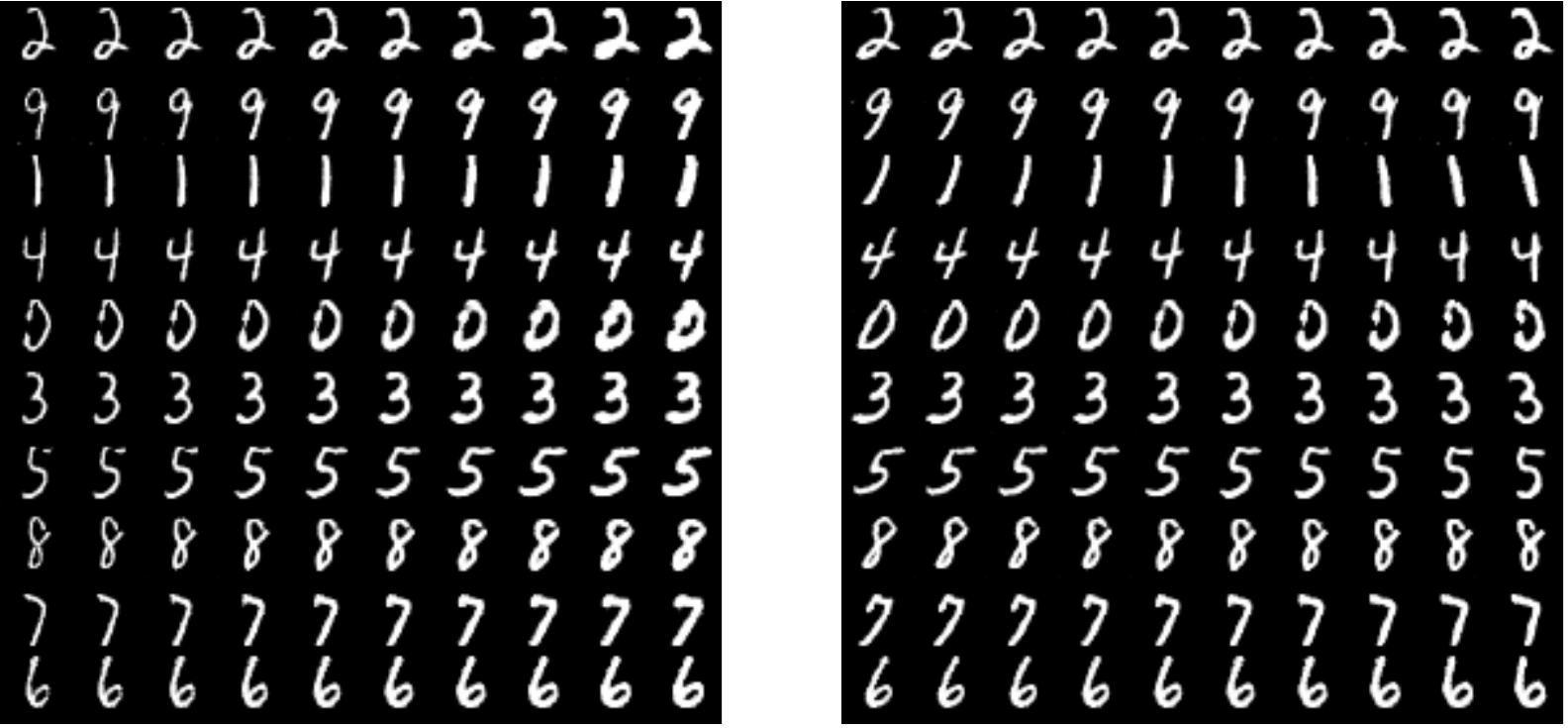}
\end{wrapfigure}
\textbf{Modeling continuous factors~}  Finally, we demonstrate that Elastic-InfoGAN does not impede modeling of continuous factors in the imbalanced setting. Specifically, one can augment the input with continuous codes (e.g., r1, r2 $\sim$ $U$ (-1, 1)) along with the existing categorical and noise vectors. In the right figure, we show the results of continuous code interpolation; we can see that each of the continuous codes largely captures a particular continuous factor (stroke width on left, digit rotation on the right).

\vspace{-6pt}
\section{Conclusion}
\vspace{-5pt}
We proposed an unsupervised generative model that better disentangles object identity as a factor of variation, without the knowledge of class imbalance. Although there are some limitations (e.g., its applicability in highly skewed data), we believe that we address an important, unexplored problem setting. Similar to how the area of supervised machine learning has evolved over time to account for class imbalance in real world data, our hope with this work is to pave the way for the evolution of unsupervised learning based methods to work well in class imbalanced data, which is inevitable if these algorithms have to be deployed in the real world.

\section*{Broader Impact}
Image datasets, particularly the ones with human faces, have a potential problem of not being diverse enough. This conceptualizes in some form of imbalance in the dataset, where, for example, a dataset of human faces might not represent faces from all ethnical communities in appropriate proportions. A popular application of GANs is to use synthetic images for data augmentations. With traditional GANs, however, it is possible that the underrepresented classes might not be modeled as accurately (mode dropping problem), thus limiting their applicability. The idea presented in this work is specifically tailored to handle such cases, by discovering both, the over and under-represented classes. This could then enable data augmentation using generated images, and help increase the proportions of underrepresented classes. 

GANs in general pose some ethical concerns, in terms of creating/altering visual content (e.g., deepfakes). Our work, which is a derivative of GAN, is no exception in that regard, as it could have some malicious applications, such as image fabrication. We do want to point out that such applications are not that straightforward with the method proposed in its current form, as our method doesn't operate directly on real images (the input to the generator is latent vectors).

\begin{ack}
This work was supported in part by  NSF IIS-1751206, IIS-1748387, IIS-1812850, IIS-1901527, IIS-2008173, AWS ML Research Award, Google Cloud Platform research credits, and Adobe Data Science Research Award.
\end{ack}

{
\bibliographystyle{plain}
\bibliography{main}
}

\newpage

\section*{Supplementary}

%
\setcounter{section}{0}
In this document, we first continue our discussion of the importance of constrastive learning of representations to disentangle object identity. We then mathematically analyze the applicability of contrastive loss in a generic class-imbalanced setting, studying how appropriate our assumptions pertaining to $L_{ntxent}$ are. We then demonstrate an application of our framework, making use of the learned inference network for nearest neighbor classification. Finally, we discuss all the implementation details, and provide detailed information about the random imbalanced splits used for different datasets.  

\section{Importance of constrastive loss}
\begin{figure*}
    \centering
    \includegraphics[width=1\textwidth]{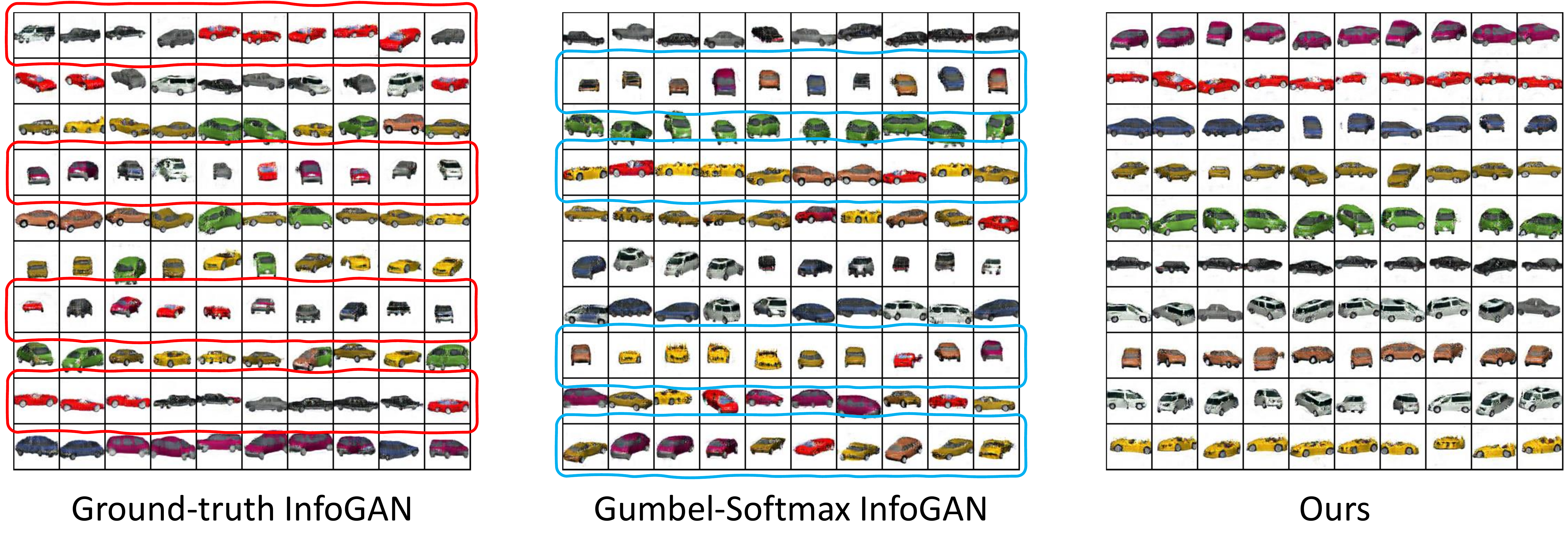}
    \caption{Image generations on a random imbalanced 3D Cars split. Baselines which could potentially capture object identity as a factor of variation, but lack $L_{ntxent}$, sometimes generate undesirable groupings, e.g.~focusing on pose/orientation rather than object identity (groups highlighted with an outline: \emph{red} for Ground-truth InfoGAN, \emph{blue} for Gumbel-Softmax InfoGAN). Our approach, for the same set of object categories, produces better groupings focusing on identity and generating cars with varying pose/orientation.}
    \label{fig:3d_cars_supp}
\end{figure*}

In this section, we continue the discussion initiated in Sec.~4.4 of the main paper, about the importance of $L_{ntxent}$. Specifically, we discussed the potential cases where Ground-truth and Gumbel-Softmax InfoGAN baselines (which could in theory produce accurate disentanglement) would perform poorly. In Fig.\ref{fig:3d_cars_supp}, we visualize these scenarios qualitatively. We highlight the cases where the mentioned baselines produce undesirable groupings; i.e., generating cars having \emph{similar} pose but \emph{different} identity for the same latent code. Our approach, for the same split of random classes, produces groupings which are much more coherent; i.e., generating cars of same identity rendered in different positions, for a latent code. 

Finally, Table \ref{table:disentangle_nmi} and \ref{table:disentangle_ent} cover the quantitative results presented in Table 1 of the main paper with error bars (standard deviation). In general, the error bars for our method are lower than the baselines, indicating more robustness is introduced via $L_{ntxent}$. The numbers are over (i) 50 different runs for MNIST (corresponding to 50 imbalanced splits), (ii) 25 runs for 3d-Cars/Chairs (refer to the main paper about the creation of 25 splits), and (iii) 5 runs over the same imbalanced split for YouTube Faces.

\section{Analysing the constrastive loss in imbalanced scenarios}
As mentioned in Sec. 3.2 of the main paper, we approximate negative samples with randomly sampled pairs $I_i$, $I_j$ ($i \neq j$) in a batch while using $L_{ntxent}$. Since we don't know the labels during training, some of these sampled pairs could turn out to be \emph{false negatives} (images belonging to the same category). In this section, we quantify the extent to which these false negatives impact the training of overall model.

Assume that the proportion of different classes in the dataset is denoted by $p_1, p_2, p_3, ..., p_k$ (where $k$ denotes the number of classes, and $p_i$ denotes $i^{th}$ class probability). Let's say that in a batch, the algorithm constructs $N$ negative pairs, where each pair is equally likely to be a false negative. Consider one such pair, $I_i$ and $I_j$: the probability that this pair will be a false negative (F.N.) is $P(F.N.) = \sum_{i=1}^{k} p_i^2$. 

Since all pairs are sampled independently, the expected fraction of false negatives, $\mathbf{E}(F.N.)$ in the batch hence becomes:
\begin{equation}
    \frac{N \times \sum_{i=1}^{k} p_i^2}{N} = \sum_{i=1}^{k} p_i^2
\end{equation}
We wish for this value to be as small as possible. As expected, the minimum value is attained when the dataset is balanced, i.e. $p_1 = p_2 ... = p_k = \frac{1}{k}$, where $\mathbf{E}(F.N.)$ becomes $\frac{1}{k}$. The maximum value, on the other hand, is attained when one class completely dominates the other classes ($p_i = 1$ for some $i$, and $p_j=  0$ ($\forall j \neq i$)), making $\mathbf{E}(F.N.) = 1$. So, as long as there are even a few prominent classes (among $k$), the expected false negatives remain considerable low. For instance, consider the randomly created imbalanced splits for MNIST (total classes = 10; detailed in Sec.\ref{sec:mnist_splits}): the average $\mathbf{E}(F.N.)$ for the 50 splits is 0.119 $\pm$ 0.008. This means that on average, ~88\% of the negative pairs will be true negative, which is only 2\% less than the ideal 90\% (when $\mathbf{E}(F.N.)$ = $\frac{1}{k}$). We further verified this hypothesis for the real-world imbalanced dataset used in this work, YouTube Faces (total classes = 40; split detailed in Sec.\ref{sec:ytf_split}). The $\mathbf{E}(F.N.)$ for this dataset is 0.0363, which means ~96\% of negative pairs are true negatives (only 1\% less than ideal scenario). 

We want to emphasize the key take away from this analysis: it is \emph{not} that the datasets explored in this work don't have sufficient imbalance (the reader can verify this by looking at the actual splits described in Sec.\ref{sec:mnist_splits} - Sec.\ref{sec:ytf_split}), but rather that the mathematical formulation of $\mathbf{E}(F.N.)$ happens to allow a large spectrum of imbalanced datasets to be applicable while using $L_{ntxent}$. 

\begin{table}
\centering
\scriptsize
\begin{tabular}{l|c|c|c|c|c}
\multicolumn{1}{c|}{}                                  & MNIST  & YTF & Cars & Chairs & ShapeNet \\ \hline
JointVAE \cite{dupont-neurips2018}                     & 0.6801 $\pm$ 0.081 & 0.4472 $\pm$ 0.027        &   0.3915 $\pm$ 0.236     &     0.4478 $\pm$ 0.212 &  0.1892 $\pm$ 0.075      \\ 
Uniform InfoGAN \cite{chen-nips16}              & 0.7765 $\pm$ 0.045 & 0.6656 $\pm$ 0.005        & 0.4990 $\pm$ 0.205  & 0.2525 $\pm$ 0.165 & 0.6382 $\pm$ 0.088    \\ 
Gumbel-softmax                & 0.8360 $\pm$ 0.048 & 0.7603 $\pm$ 0.014        & 0.4003 $\pm$ 0.309  & 0.2356 $\pm$ 0.134 & 0.6031 $\pm$ 0.153   \\ 
Gumbel-softmax + pos-$L_{ntxent}$        & 0.8781 $\pm$ 0.061 & 0.7647 $\pm$ 0.011        & 0.5818 $\pm$ 0.127  & 0.4536 $\pm$ 0.172 & 0.7236 $\pm$ 0.062    \\ 
Elastic-InfoGAN (Ours)       & \textbf{0.8893 $\pm$ 0.044} & \textbf{0.7923 $\pm$ 0.004}         & \textbf{0.8504 $\pm$ 0.068}  & \textbf{0.6499 $\pm$ 0.082}  & \textbf{0.7900 $\pm$ 0.077}   \\ \hline
Ground-truth InfoGAN          & 0.7827 $\pm$ 0.049 & 0.6941 $\pm$ 0.031        & 0.4512 $\pm$ 0.276  & 0.1738 $\pm$ 0.142  & 0.5492 $\pm$ 0.152  \\ 
Ground-truth InfoGAN + $L_{ntxent}$ & 0.8008 $\pm$ 0.052 & 0.7421 $\pm$ 0.023        & 0.7841 $\pm$ 0.092  & 0.5915 $\pm$ 0.152  & 0.5314 $\pm$ 0.105 \\ \hline
\end{tabular}
\vspace{2pt}
\caption{Distentanglement quality measured by NMI ($\uparrow$ better). The first five methods do not have knowledge of the ground-truth distribution, while the last two methods do. Our model outperforms the baselines with/without the knowledge of ground-truth class distribution, for all datasets, with relatively low error-bar}
\label{table:disentangle_nmi}
\end{table}

\begin{table}[t]
\centering
\scriptsize
\begin{tabular}{l|c|c|c|c|c}

\multicolumn{1}{c|}{}        & MNIST  & YTF & Cars & Chairs & ShapeNet \\ \hline
JointVAE \cite{dupont-neurips2018}                     & 0.7006 $\pm$ 0.134 & 1.7173 $\pm$ 0.022        &  1.0818  $\pm$ 0.526     &   1.9090 $\pm$ 0.438  & 1.0113 $\pm$ 0.064      \\ 
Uniform InfoGAN \cite{chen-nips16}              & 0.4569 $\pm$ 0.096 & 1.0312 $\pm$ 0.009        & 1.1075 $\pm$ 0.483  & 1.6626 $\pm$ 0.368  & 0.5318 $\pm$ 0.141  \\ 
Gumbel-softmax                & 0.3260 $\pm$ 0.101 & 0.7573 $\pm$ 0.014        & 1.3183 $\pm$ 0.702  & 1.6960 $\pm$ 0.305  & 0.6188 $\pm$ 0.241  \\ 
Gumbel-softmax + pos-$L{ntxent}$        & 0.2347 $\pm$ 0.101 & 0.7188 $\pm$ 0.006        & 0.9188 $\pm$ 0.247  & 1.2088 $\pm$ 0.417  & 0.3973 $\pm$ 0.069  \\
Elastic-InfoGAN (Ours)        & \textbf{0.2130 $\pm$ 0.088} & \textbf{0.6358 $\pm$ 0.015}        & \textbf{0.3026 $\pm$ 0.147}  & \textbf{0.7651 $\pm$ 0.180} & \textbf{0.2972 $\pm$ 0.104}    \\ \hline
Ground-truth InfoGAN          & 0.4196 $\pm$ 0.097 & 0.9611 $\pm$ 0.009       & 1.1907 $\pm$ 0.646  & 1.8369 $\pm$ 0.318 & 0.6730 $\pm$ 0.249 \\ 
Ground-truth InfoGAN + $L_{ntxent}$ & 0.3694 $\pm$ 0.094 & 0.7672 $\pm$ 0.007        & 0.4373 $\pm$ 0.191  & 0.8851 $\pm$ 0.331  & 0.7162 $\pm$ 0.172  \\ \hline
\end{tabular}
\vspace{2pt}
\caption{Distentanglement quality measured by ENT ($\downarrow$ better).}
\label{table:disentangle_ent}
\end{table}

\section{k-Nearest neighbour classification}
We discussed in Sec. 1 and 3.2 of the main paper that disentangling object identity from other factors can have downstream applications. One such application can be demonstrated via the trained inference network ($Q$), where it can be used to extract features pertaining to object identity of any real image $x$. We present results in scenarios of class-imbalance, where it becomes difficult to perform grouping based on object identity.

Table~\ref{table:knn} summarizes the performance of different methods for the task of nearest neighbours classification. 
Randomly initialized refers to using the same inference network architecture with random weights. In particular, we create a 80/20 (train/test) split of real images, and report classification accuracy when images from the test split are used as queries. We see that our method consistently achieves superior performance across the four datasets. Note that the performance on YTF is remarkably good because images for each class are almost identical, so the embedding for image for each class will be very close to each other, regardless of the encoder. 

\begin{figure*}
    \centering
    \includegraphics[width=1\textwidth]{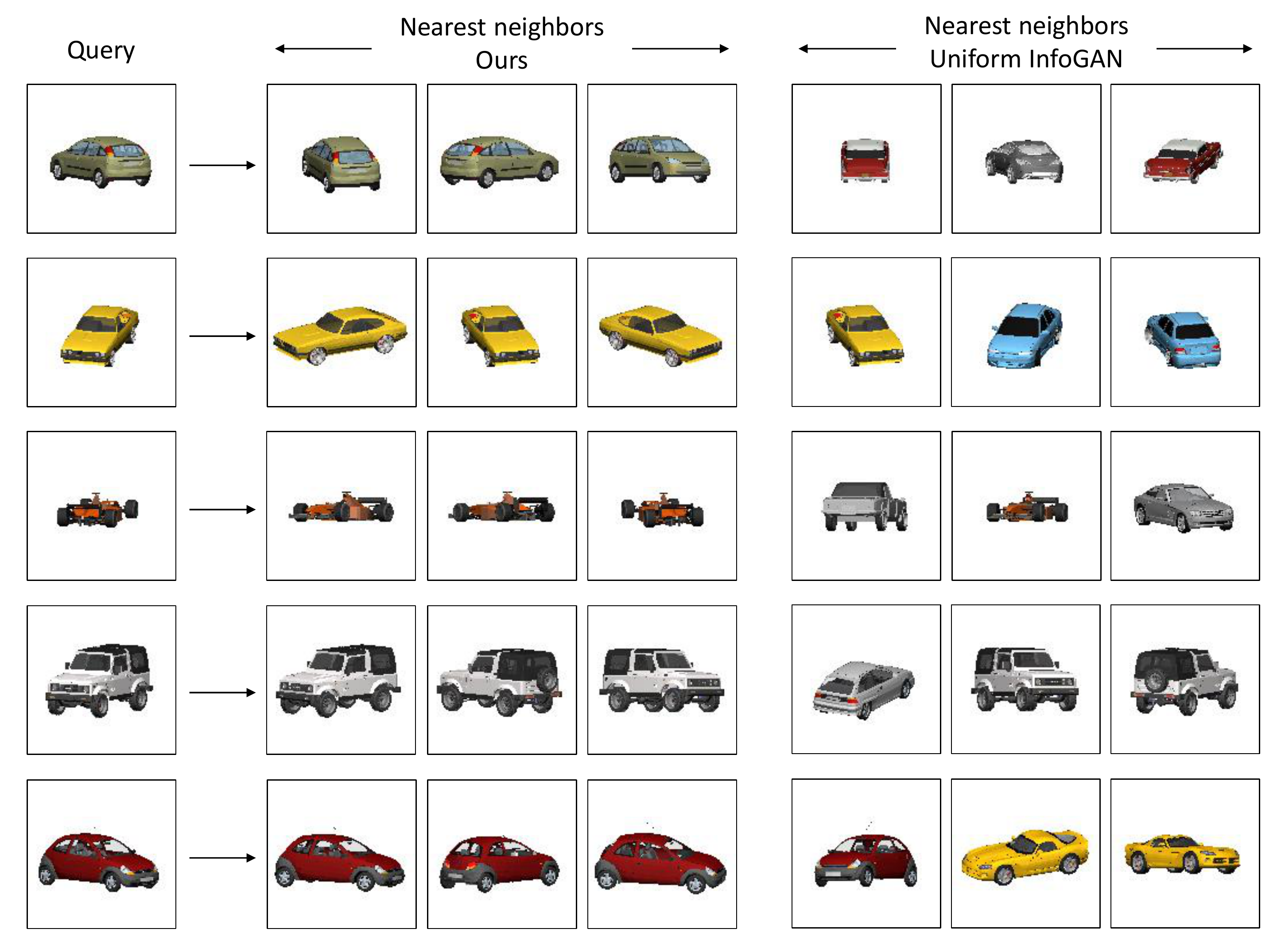}
    \caption{Nearest neighbors obtained using our method vs. uniform InfoGAN. We see that our method retrieves diverse cars (in terms of azimuth, pose etc.) belonging to the same category. Neighbors extracted using Uniform InfoGAN, on the other hand, suffer in terms of categorical consistency.}
    \label{fig:nn_res}
\end{figure*}

Fig. \ref{fig:nn_res} illustrates some sample queries, and the corresponding nearest neighbors retrieved using uniform InfoGAN and our method. The neighbors extracted using uniform InfoGAN suffer inconsistency, where sometimes the nearest neighbor has similarities in pose (3rd row), rough color (4th row) etc. Our method, on the other hand, has much more success in retrieving images belonging to the same category. Furthermore, note the variations among the query image and the extracted images for our method. It is not possible to cover all these differences in pose, azimuth through simple transformations ($\delta$) that we use in $L_{ntxent}$. This is an indication that our method is not simply memorizing the transformations that we introduce, and is actually learning representations which focus on object identity in general, leading to superior performance in 1NN classification.

\vspace{10pt}
\begin{table}[]
\centering
\scriptsize
\begin{tabular}{l|c|c|c|c}
                     & MNIST  & YTF & 3D Cars & 3D Chairs \\ \hline
Randomly initialized & 0.4865 $\pm$ 0.009 & 0.9541 $\pm$ 0.003   & 0.3140 $\pm$ 0.055  & 0.2942 $\pm$ 0.038   \\
Uniform InfoGAN \cite{chen-nips16}     & 0.8665 $\pm$ 0.014 & 0.9581 $\pm$ 0.026   & 0.7700 $\pm$ 0.074  & 0.6438 $\pm$ 0.109   \\
Gumbel-Softmax       & 0.9113 $\pm$ 0.017 & 0.9812 $\pm$ 0.013   & 0.6726 $\pm$ 0.073  & 0.6241 $\pm$ 0.119   \\
Elastic-InfoGAN (Ours)                 & \textbf{0.9655 $\pm$ 0.007} & \textbf{0.9985 $\pm$ 0.018}   & \textbf{0.9852 $\pm$ 0.005}  & \textbf{0.9515 $\pm$ 0.020}   \\ \hline
\end{tabular}
\vspace{1pt}
\caption{1NN classification accuracy (\%) of different baselines. By learning to better disentangle object identity from other factors, our method can infer much better representations needed for the task of nearest neighbor classification.} 
\label{table:knn}
\end{table}

\section{Implementation details (continued)}
\subsection{MNIST}
For MNIST, we operate on the original 28x28 images, with a 10-dimensional categorical code to represent the 10 digit categories, and 62 noise variables sampled from a normal distribution. We follow the exact architecture as described in InfoGAN~\cite{chen-nips16}: The generator network $G$ takes as input a $64$ dimensional noise vector $z \sim \mathcal{N}(0, 1)$ and 10 dimensional samples from a Gumbel-Softmax distribution. The discriminator $D$ and the latent code prediction network $Q$ share most of the layers except the final fully connected layers. 

The pre-trained classification architecture used for evaluation for MNIST consists of 2 Conv + 2 FC layers, with max pool and ReLU after every convolutional layer.

\subsection{3D Cars and Chairs}
We follow identical steps for 3D Cars and Chairs. For 3D Cars, we follow the procedure explained in the main paper to select 10 categories. We resize all the renderings to 64x64 resolution, and use a 10 dimensional categorical code to represent 10 object identities, and use 100 noise variables to capture other variations (pose/viewpoints etc.). 

Our architecture is based on the one proposed in StackGANv2~\cite{zhang-tpami2018}, where we use its 1-stage version for generating 64x64x3 resolution images. There is an initial fully connected layer which maps the input (concatenation of $z$ and $c$) to an intermediate feature representation. A series of a combination of upsampling + convolutional layers (interleaved with batch normalization and Gated Linear Units) increase the spatial resolution of the feature representation, starting from 1024 (feature size: 4 x 4 x 1024) channels to 64 (feature size: 64 x 64 x 64) channels. A convolutional network transforms the feature representation into a 3 channel output, while maintaining the spatial resolution, which serves as the fake image. The discriminator network consists of 4 convolutional layers interleaved with batch normalization and leaky ReLU layers, which serve as the common layers for both the $D$ and $Q$ networks. After that, $D$ has one non-shared convolutional layer, which maps the feature representation into a scalar value reflecting the real/fake score. For $Q$, we have a pair of non-shared convolutional layers which map the feature representation into a 10 dimensional latent code prediction.

The classifier used for evaluating results on 3D Cars/Chairs is a ResNet-50 network, trained on the complete data with a 80/20 train/validation split (different pre-trained networks for different sets of 10 classes chosen).

\subsection{YouTube-Faces}
For YouTube-Faces, we crop the faces using the provided bounding box annotations, and then resize them to 64x64 resolution, and use a 40-dimensional categorical code to represent 40 face identities (first 40 categories sorted in alphabetical manner, as done in~\cite{shah-arxiv18}), and 100 noise variables. 

The architecture for the Generator/Discriminator is very much similar to that used for 3D Cars/Chairs, except that we have one more stage, which takes in the 64 x 64 x 64 resolution intermediate features and translates that into another fake image. We apply our losses on images from both the stages, and the images generated from the second stage are used for evaluation purposes.

The classifier used for evaluating results on YouTube-Faces is also ResNet-50 network (similar to the one used for 3D Cars/Chairs), but we pretrain it on VGGFace2, before fine-tuning on YouTube-Faces.

\subsection{Training details}
We employ a similar way of training the generative and discriminative modules as described in ~\cite{chen-nips16}. We first update the discriminator based on the real/fake adversarial loss. In the next step, after computing the remaining losses (mutual information + $L_{ntxent}$), we update the generator ($G$) + latent code predictor ($Q$) + latent distribution parameters ($p_{i_s}$) at once. Our optimization process alternates between these two phases. $L_{ntxent}$ is computed on features obtained from penultimate layer, after the leaky ReLU activation. One good reason to not use ReLU instead is that it will produce bias for cosine similarity distance, since all the feature values will be positive. We use Adam optimizer, with a learning rate of 0.0002. For MNIST, we train all baselines for 200 epochs, with a batch size of 64. For 3D Cars/Chairs, we train for 600 epochs, with a batch size of 50. For YouTube-Faces, we train until convergence, as measured via qualitative realism of the generated images, using a batch size of 50. $\lambda_2$ in $L_{final}$ is set to 10 to balance the magnitude of the different loss terms. $\tau = 0.1$ is used for sampling from Gumbel-Softmax, which results in samples having very low entropy (very close to one hot vectors from a categorical distribution).

\textbf{JointVAE details:} We use the KL term for both continuous as well as discrete variables, to follow the standard normal ($\mathcal{N}(0,1)$) and uniform categorical distribution (Cat$(p = 1/k)$), respectively. We use uniform categorical because of the unsupervised nature of the problem (L53-5, 83-4). We use the same weight ($\beta$) for both KL loss terms (similar to JointVAE paper), the value of which was first decided empirically based on image reconstruction quality - we observed that a value in 100s (e.g. 100-300) resulted in poor reconstruction quality. After that, we report the results for best performing model by ablating $\beta_{cont}$ and $\beta_{disc}$ from the set \{10, 20, 30, 40, 50\}.

Finally, one behavior we observe is that if the random initialization of class probabilities is too skewed (only few classes have high probability values), then it becomes very difficult for them to get optimized to the ideal state. We hence initialize them with the uniform distribution, which makes training much more stable.

Experiments on MNIST, which involve running 50 versions, can be done in parallel across 4 NVIDIA Tesla V100 GPUs (16 GB RAM). On each of them, around 6  (out of 50) can run in parallel. In this manner, it takes about 8 hours to complete the training on 50 MNIST imbalanced splits. For 3D Cars/Chairs, it takes about 4-5 hours if proper parallelism is employed on the same system. For YTF, it takes about 4 hours to run 5 versions on the same imbalanced split (while running in parallel). 
\section{Ground truth class imbalance}
Here we describe the exact class imbalance used in our experiments. For MNIST, we include below the 50 random imbalances created. For 3D Cars/Chairs, we first describe the class ids of the randomly chosen 10 categories for all the 5 sets, then we describe the 5 random imbalanced splits used on each of these sets. For YouTube-Faces, we include the true ground truth class imbalance in the first 40 categories. The imbalances reflect the class frequency.

\subsection{MNIST}\label{sec:mnist_splits}
\begin{itemize}
\fontsize{8}{12}\selectfont
\vspace{-5pt}\item 0.147, 0.037, 0.033, 0.143, 0.136, 0.114, 0.057, 0.112, 0.143, 0.078
\vspace{-5pt}\item 0.061, 0.152, 0.025, 0.19, 0.12, 0.036, 0.092, 0.185, 0.075, 0.064
\vspace{-5pt}\item 0.173, 0.09, 0.109, 0.145, 0.056, 0.114, 0.075, 0.03, 0.093, 0.116
\vspace{-5pt}\item 0.079, 0.061, 0.033, 0.139, 0.145, 0.135, 0.057, 0.062, 0.169, 0.121
\vspace{-5pt}\item 0.053, 0.028, 0.111, 0.142, 0.13, 0.121, 0.107, 0.066, 0.125, 0.118
\vspace{-5pt}\item 0.072, 0.148, 0.092, 0.081, 0.119, 0.172, 0.05, 0.109, 0.085, 0.073
\vspace{-5pt}\item 0.084, 0.143, 0.07, 0.082, 0.059, 0.163, 0.156, 0.063, 0.074, 0.105
\vspace{-5pt}\item 0.062, 0.073, 0.065, 0.183, 0.099, 0.08, 0.05, 0.16, 0.052, 0.177
\vspace{-5pt}\item 0.139, 0.113, 0.074, 0.06, 0.068, 0.133, 0.142, 0.13, 0.112, 0.03
\vspace{-5pt}\item 0.046, 0.128, 0.059, 0.112, 0.135, 0.164, 0.142, 0.125, 0.051, 0.037
\vspace{-5pt}\item 0.107, 0.057, 0.154, 0.122, 0.05, 0.111, 0.032, 0.044, 0.136, 0.187
\vspace{-5pt}\item 0.129, 0.1, 0.039, 0.112, 0.119, 0.095, 0.047, 0.14, 0.156, 0.064
\vspace{-5pt}\item 0.146, 0.08, 0.06, 0.072, 0.051, 0.119, 0.176, 0.11, 0.158, 0.028
\vspace{-5pt}\item 0.035, 0.051, 0.112, 0.143, 0.033, 0.165, 0.082, 0.165, 0.054, 0.161
\vspace{-5pt}\item 0.041, 0.1, 0.073, 0.054, 0.155, 0.117, 0.091, 0.124, 0.142, 0.104
\vspace{-5pt}\item 0.052, 0.139, 0.128, 0.133, 0.104, 0.107, 0.058, 0.137, 0.036, 0.107
\vspace{-5pt}\item 0.055, 0.138, 0.059, 0.074, 0.08, 0.135, 0.085, 0.064, 0.172, 0.139
\vspace{-5pt}\item 0.141, 0.156, 0.119, 0.062, 0.08, 0.022, 0.043, 0.159, 0.101, 0.118
\vspace{-5pt}\item 0.11, 0.088, 0.033, 0.062, 0.089, 0.176, 0.161, 0.105, 0.144, 0.032
\vspace{-5pt}\item 0.157, 0.111, 0.125, 0.099, 0.036, 0.119, 0.036, 0.05, 0.147, 0.121
\vspace{-5pt}\item 0.119, 0.121, 0.117, 0.152, 0.026, 0.174, 0.027, 0.065, 0.151, 0.049
\vspace{-5pt}\item 0.057, 0.07, 0.134, 0.118, 0.058, 0.185, 0.07, 0.13, 0.116, 0.063
\vspace{-5pt}\item 0.102, 0.082, 0.135, 0.046, 0.128, 0.106, 0.116, 0.085, 0.133, 0.066
\vspace{-5pt}\item 0.057, 0.193, 0.2, 0.123, 0.022, 0.154, 0.115, 0.025, 0.065, 0.047
\vspace{-5pt}\item 0.056, 0.196, 0.168, 0.052, 0.116, 0.062, 0.099, 0.133, 0.065, 0.053
\vspace{-5pt}\item 0.04, 0.022, 0.2, 0.194, 0.038, 0.033, 0.161, 0.097, 0.159, 0.056
\vspace{-5pt}\item 0.04, 0.036, 0.119, 0.204, 0.16, 0.103, 0.089, 0.061, 0.136, 0.052
\vspace{-5pt}\item 0.112, 0.189, 0.145, 0.163, 0.113, 0.031, 0.028, 0.062, 0.045, 0.112
\vspace{-5pt}\item 0.071, 0.099, 0.113, 0.175, 0.082, 0.068, 0.03, 0.066, 0.133, 0.164
\vspace{-5pt}\item 0.134, 0.074, 0.111, 0.091, 0.051, 0.119, 0.044, 0.085, 0.144, 0.148
\vspace{-5pt}\item 0.103, 0.126, 0.084, 0.117, 0.084, 0.127, 0.131, 0.092, 0.117, 0.019
\vspace{-5pt}\item 0.096, 0.121, 0.026, 0.046, 0.043, 0.124, 0.165, 0.04, 0.127, 0.213
\vspace{-5pt}\item 0.117, 0.115, 0.125, 0.128, 0.081, 0.103, 0.073, 0.044, 0.137, 0.077
\vspace{-5pt}\item 0.037, 0.021, 0.143, 0.165, 0.075, 0.111, 0.028, 0.132, 0.134, 0.154
\vspace{-5pt}\item 0.154, 0.049, 0.128, 0.089, 0.082, 0.072, 0.034, 0.138, 0.108, 0.146
\vspace{-5pt}\item 0.078, 0.141, 0.084, 0.139, 0.085, 0.062, 0.035, 0.174, 0.15, 0.053
\vspace{-5pt}\item 0.112, 0.112, 0.128, 0.112, 0.107, 0.142, 0.032, 0.142, 0.063, 0.049
\vspace{-5pt}\item 0.084, 0.091, 0.128, 0.129, 0.045, 0.105, 0.05, 0.091, 0.089, 0.188
\vspace{-5pt}\item 0.062, 0.136, 0.112, 0.153, 0.091, 0.046, 0.089, 0.03, 0.161, 0.12
\vspace{-5pt}\item 0.143, 0.1, 0.046, 0.166, 0.107, 0.191, 0.026, 0.078, 0.097, 0.047
\vspace{-5pt}\item 0.077, 0.174, 0.05, 0.098, 0.028, 0.173, 0.067, 0.106, 0.096, 0.13
\vspace{-5pt}\item 0.105, 0.022, 0.183, 0.056, 0.045, 0.103, 0.081, 0.135, 0.119, 0.149
\vspace{-5pt}\item 0.083, 0.127, 0.126, 0.028, 0.209, 0.03, 0.066, 0.125, 0.1, 0.107
\vspace{-5pt}\item 0.138, 0.142, 0.074, 0.091, 0.103, 0.067, 0.12, 0.04, 0.1, 0.124
\vspace{-5pt}\item 0.058, 0.039, 0.088, 0.113, 0.093, 0.055, 0.162, 0.069, 0.168, 0.155
\vspace{-5pt}\item 0.02, 0.162, 0.133, 0.138, 0.137, 0.051, 0.069, 0.032, 0.118, 0.14
\vspace{-5pt}\item 0.071, 0.046, 0.134, 0.119, 0.159, 0.057, 0.039, 0.135, 0.057, 0.184
\end{itemize}

\subsection{3D Cars}
List of class ids for different (total 5) sets of randomly chosen classes:
\begin{itemize}
\fontsize{8}{12}\selectfont
    \item 009, 002, 004, 007, 001, 025, 026, 024, 043, 023
    \vspace{-5pt}
    \item 096, 118, 040, 052, 024, 046, 123, 187, 150, 072
    \vspace{-5pt}
    \item 112, 019, 030, 037, 069, 056, 161, 193, 190, 061
    \vspace{-5pt}
    \item 038, 111, 104, 159, 035, 037, 086, 043, 173, 196
    \vspace{-5pt}
    \item 113, 009, 031, 016, 022, 078, 083, 060, 098, 100
\end{itemize}

Imbalances splits applied on each of these sets:
\begin{itemize}
\fontsize{8}{12}\selectfont
    \item 0.141, 0.116, 0.128, 0.077, 0.104    \vspace{-5pt}
    \item 0.035, 0.137, 0.027, 0.068, 0.175    \vspace{-5pt}
    \item 0.081, 0.076, 0.117, 0.109, 0.079    \vspace{-5pt}
    \item 0.134, 0.108, 0.048, 0.143, 0.107    \vspace{-5pt}
    \item 0.033, 0.111, 0.155, 0.160, 0.167
\end{itemize}

\subsection{3D Chairs}
List of class ids for different (total 5) sets of randomly chosen classes:
\begin{itemize}
\fontsize{8}{12}\selectfont
    \item 0965, 0960, 0710, 0045, 1332, 0996, 1074, 0236, 0098, 1196
    \vspace{-5pt}
    \item 0241, 0307, 0091, 1071, 1317, 0104, 1098, 1064, 0158, 0784
    \vspace{-5pt}
    \item 0565, 0326, 0892, 0308, 0858, 1212, 0802, 0236, 0257, 0749
    \vspace{-5pt}
    \item 0241, 0574, 0864, 0401, 1372, 1032, 1101, 0439, 0528, 0264
    \vspace{-5pt}
    \item 0561, 0334, 1036, 0724, 1314, 0766, 0572, 0840, 1338, 0991
\end{itemize}

Imbalances splits applied on each of these sets:
\begin{itemize}
\fontsize{8}{12}\selectfont
    \item 0.190, 0.040, 0.107, 0.170, 0.101    \vspace{-5pt}
    \item 0.164, 0.204, 0.060, 0.164, 0.055    \vspace{-5pt}
    \item 0.084, 0.119, 0.188, 0.067, 0.070    \vspace{-5pt}
    \item 0.035, 0.178, 0.130, 0.102, 0.173    \vspace{-5pt}
    \item 0.044, 0.060, 0.191, 0.100, 0.022
\end{itemize}

\subsection{YouTube-Faces}\label{sec:ytf_split}
\begin{itemize}
\fontsize{8}{12}\selectfont
\vspace{-5pt}\item 0.019, 0.013, 0.024, 0.020, 0.028, 0.022, 0.053, 0.010, 0.062, 0.031, 0.037, 0.005, 0.011, 0.027, 0.034, 0.033, 0.009, 0.006, 0.011, 0.016, 0.024, 0.047, 0.028, 0.069, 0.012, 0.006, 0.024, 0.005, 0.006, 0.024, 0.005, 0.037, 0.028, 0.056, 0.059, 0.026, 0.008, 0.006, 0.028, 0.028
\end{itemize}

\subsection{ShapeNet}\label{sec:shapenet_split}
\begin{itemize}
\fontsize{8}{12}\selectfont
\vspace{-5pt}\item 0.1851, 0.1481, 0.1111, 0.2592, 0.2962
\end{itemize}
Results are reported averaged over 5 different runs on this imbalance split.

\end{document}